\documentclass{article}
\usepackage[english]{babel}
\usepackage[letterpaper,top=2cm,bottom=2cm,left=3cm,right=3cm,marginparwidth=1.75cm]{geometry}
\usepackage{amsmath, amssymb}
\usepackage{graphicx}
\usepackage{booktabs}
\usepackage{float}
\usepackage[table]{xcolor}
\definecolor{brown}{RGB}{150,75,0}
\usepackage[colorlinks=true, allcolors=blue]{hyperref}
\usepackage{makecell}
\usepackage{lmodern}

\title{\textbf{Attention via Synaptic Plasticity is All You Need:\\
A Biologically Inspired Spiking Neuromorphic Transformer}}

\author{%
\begin{tabular}{c}
Kallol~Mondal$^{1,2}$, Ankush~Kumar$^{2,*}$\\[4pt]
\small $^{1}$Department of Electronics and Communication Engineering,\\
\small National Institute of Technology Allahabad, Prayagraj, 211004, India\\[6pt]
\small $^{2}$Centre for Nanotechnology, Indian Institute of Technology Roorkee,\\
\small Roorkee, Uttarakhand, 247667, India\\[6pt]
\small *Corresponding author email: ankush.kumar@nt.iitr.ac.in
\end{tabular}
}

\date{}  % removes the date

\begin{document}
\maketitle

\begin{abstract}
Attention is the brain's ability to selectively focus on a few specific aspects while ignoring irrelevant ones. This biological principle inspired the attention mechanism in modern Transformers. Transformers now underpin large language models (LLMs) such as GPT, but at the cost of massive training and inference energy, leading to a large carbon footprint. While brain attention emerges from neural circuits, Transformer attention relies on dot-product similarity to weight elements in the input sequence. Neuromorphic computing, especially spiking neural networks (SNNs), offers a brain-inspired path to energy-efficient intelligence. Despite recent work on attention-based spiking Transformers, the core attention layer remains non-neuromorphic. Current spiking attention (i) relies on dot-product or element-wise similarity suited to floating-point operations, not event-driven spikes; (ii) keeps attention matrices that suffer from the von Neumann bottleneck, limiting in-memory computing; and (iii) still diverges from brain-like computation. To address these issues, we propose the Spiking STDP Transformer (S$^{2}$TDPT), a neuromorphic Transformer that implements self-attention through spike-timing-dependent plasticity (STDP), embedding query--key correlations in synaptic weights. STDP, a core mechanism of memory and learning in the brain and widely studied in neuromorphic devices, naturally enables in-memory computing and supports non-von Neumann hardware. On CIFAR-10 and CIFAR-100, our model achieves 94.35\% and 78.08\% accuracy with only four timesteps and 0.49 mJ on CIFAR-100, an 88.47\% energy reduction compared to a standard ANN Transformer. Grad-CAM shows that the model attends to semantically relevant regions, enhancing interpretability. Overall, S$^{2}$TDPT illustrates how biologically inspired attention can yield energy-efficient, hardware-friendly, and explainable neuromorphic models.
\end{abstract}
\section{Introduction}
Attention is the central cognitive activity to focus on selected information from the noise of thoughts. When we observe something through our senses, we might be focusing only on a particular object while ignoring a very large part of the background. When we fail to direct attention toward a specific thought, we lose regulatory control over our thinking leading to confusion and unstructured mental activity. On the other hand right amount and placement of focus can transform an unstructured mental state into a calm mind required for efficient decision making \cite{Gita, Gita2}. 
Inspired by the human brain, the attention mechanism has also been implemented in machine learning, more specifically in Transformer based artificial neural networks. Modern artificial intelligence experienced a breakthrough with the introduction of the Transformer architecture in "Attention Is All You Need" \cite{r1}. Attention-based Transformers have rapidly become the backbone of natural language processing (NLP), as demonstrated by breakthrough large language models (LLMs) such as GPT, Claude, and Gemini. They now have wide applications in diverse domains, including image classification \cite{r10}, object detection \cite{r11}, and image segmentation \cite{r12}. Their exceptional ability to model long-range dependencies and learn powerful representations has enabled state-of-the-art performance across many tasks \cite{r1,r10,r12}. However, this progress has come at a considerable cost: massive parameter counts, extensive training procedures, and overwhelming energy demands. In contrast, the brain performs these cognitive activities in several orders of lower energy  exploring the adaptability, diversity, and plasticity of spikes, neurons, and synapses. Thus, inspired by the brain neuromorphic computing mimic the discrete spike-based communication of biological neurons, synapses and enable sparse, event-driven processing in  contrast to traditional neural networks. As the third generation of neural network models \cite{maass1997}, SNNs offer energy-efficient computation while maintaining rich computational capabilities \cite{pfeiffer2018deep}. A growing body of research now explores the intersection of these two paradigms, leading to Spiking Transformers-hybrid architectures that merge the representational power of Transformers with the efficiency of SNNs. By replacing continuous activations with discrete spikes, these models retain the global attention mechanism while dramatically improving energy efficiency. On standard benchmarks, they often achieve performance comparable to, or even exceeding, conventional artificial neural networks, validating this integration of symbolic and spiking computation \cite{Spikeformer,Spikformer, Spikingformer, Additon_only_Spiking_Transformer,Spiking_Vit_with_Saccadic_Attention, Spike-driven_Transformer_V2, QKFormer, ReverB-SNN,STEP}. These works aim to leverage the event-driven computation of SNNs while preserving the powerful sequence modeling ability of Transformers. The self-attention mechanism remains at the heart of this integration. In the Vanilla self-attention(VSA) framework \cite{r1}, three fundamental components are utilized: Query (Q), Key (K), and Value (V). As depicted in Eq.~\ref{e2}, the procedure first computes the similarity between Q and K through a floating-point dot-product operation, generating an attention score matrix. This matrix is subsequently passed through a softmax normalization, which applies exponential and division operations to generate the attention map \cite{r1}. The resulting map determines how different elements of V are weighted, allowing the model to selectively focus on the most relevant features. Given an input sequence of embeddings \( X = [x_1, x_2, \ldots, x_n] \),  these embeddings are linearly transformed into the query, key, and value spaces using learnable weight matrices:
\begin{equation}
Q = XW_Q,\quad K = XW_K,\quad V = XW_V,
\end{equation}
where \( W_Q \), \( W_K \), and \( W_V \) denote the trainable projection matrices for Q, K, and V, respectively.  
\begin{equation}
\text{Attention}(Q, K, V) = \text{softmax}\left(\frac{QK^{T}}{\sqrt{d_k}}\right)V,
\label{e2}
\end{equation}
where \( d_k \) represents the dimension of the key vectors. But The floating-point matrix multiplication and the softmax function, which involve exponential and division operations, are not inherently compatible with the characteristics of SNNs \cite{QKFormer}. The two pioneering works in this direction are Spikformer \cite{Spikformer} and Spikeformer \cite{Spikeformer}, both of which introduced Transformer-inspired architectures for spiking neural networks. Even though Spikeformer\cite{Spikeformer} introduces spiking activation functions in the feedforward blocks, it continues to rely on several non-spiking computations, notably floating-point multiplication, division, and exponential operations. To address this limitation, Spikformer\cite{Spikformer} introduces a novel Spiking Self-Attention (SSA) framework, specifically designed to align with the spike-driven processing paradigm. In this approach, the query (\(Q\)), key (\(K\)), and value (\(V\)) representations are first obtained through trainable linear projections. These continuous-valued representations are then transformed into spike sequences by applying batch normalization followed by binary spiking neuron layers:
\begin{equation}
Q = SN_{Q}\bigl(BN(XW_{Q})\bigr),
\end{equation}
\begin{equation}
K = SN_{K}\bigl(BN(XW_{K})\bigr),
\end{equation}
\begin{equation}
V = SN_{V}\bigl(BN(XW_{V})\bigr),
\end{equation}
where \( BN(\cdot) \) denotes the batch normalization operation and \( SN(\cdot) \) represents the binary spiking neuron layer. Unlike the standard softmax-based attention, the attention scores in SSA are derived from binary spike representations of \( Q \) and \( K \), which contain only 0s and 1s. This design enables the replacement of floating-point matrix multiplications with addition-based computations, making the mechanism more compatible with SNN hardware. The final SSA operation is formulated as:
\begin{equation}
\text{SSA}(Q, K, V) = SN\bigl(QK^{T} V \cdot s\bigr).
\end{equation}
Here, the scaling parameter \( s \) adjusts the magnitude of the result from the matrix product ensuring that the attention values remain stable and do not become too high during processing \cite{Additon_only_Spiking_Transformer}. All subsequent operations within the SSA module rely solely on addition, maintaining consistency with the event-driven and energy-efficient nature of SNNs. Although Spikformer\cite{Spikformer} achieves state-of-the-art performance on several datasets, its residual connection structure still involves non-spiking computations \cite{Spikingformer}. To address this limitation, Spikingformer \cite{Spikingformer} adopts the SSA module from Spikformer \cite{Spikformer} and further incorporates spike-driven residual blocks to build a fully spiking Transformer architecture. 
While Spikformer and Spikingformer successfully adapted self-attention to the spiking domain, they inherited the quadratic complexity bottleneck of traditional transformers with respect to sequence length \cite{QKFormer}. QKFormer addresses this fundamental limitation by introducing a novel Q-K attention mechanism that reduces the traditional three-component attention (Query-Key-Value) to a two-component formulation, eliminating the Value matrix entirely \cite{QKFormer}. This architectural simplification yields multiple benefits: reduced synaptic operations, lower memory requirements, and linear computational complexity with respect to either token number or feature dimension. The Q-K attention operates through two complementary variants: Q-K Token Attention (QKTA) and Q-K Channel Attention (QKCA). QKTA performs row summation on the Query matrix followed by column masking operations on the Key matrix, achieving O(N) or O(D) complexity compared to O(N²D) in vanilla self-attention~\cite{QKFormer}.

Despite recent architectural advances in reducing computational complexity and enabling hierarchical designs, a fundamental challenge remains: binary spike activations inherently constrain the expressive power of attention computations due to their limited information capacity \cite{Additon_only_Spiking_Transformer}. The Accurate Addition-Only Spiking Self-Attention (A$^{2}$OS$^{2}$A) framework provides a rigorous information-theoretic foundation for analyzing these representational constraints \cite{Additon_only_Spiking_Transformer}. Through an entropy-based analysis, the study quantifies the information capacity gap between binary spike layers and real-valued layers. This theoretical insight establishes that employing binary spike activations across all Query, Key, and Value components sacrifices critical representational fidelity, thereby motivating the need for alternative neuron designs that selectively allocate precision where it is most beneficial. To address this information bottleneck, A$^{2}$OS$^{2}$A introduces a heterogeneous neuron strategy that balances information capacity with computational efficiency through selective precision allocation. The approach employs three distinct neuron types within the attention mechanism: Binary Spiking Neurons for Query matrices ($Q$), ReLU activations for Key matrices ($K$), Ternary Spiking Neurons for Value matrices ($V$) while maintaining addition-only operations. This hybrid neuron design eliminates both softmax and scaling operations. The non-negative attention map ($QK^{T}$) emerges naturally from the non-negative $Q$ and $K$, thus removing the necessity for softmax normalization. Similarly, the unbounded nature of $K$ obviates the need for scaling factors. The matrix multiplication ($QK^{T}$) is replaced by an addition-based operation due to the binary nature of $Q$, and the subsequent computation ($(QK^{T})V$) also becomes addition-only since $V$ is ternary. This formulation achieves a truly "addition-only" attention mechanism while preserving sufficient representational capacity through strategic precision distribution across different attention components\cite{Additon_only_Spiking_Transformer}.

While the Accurate Addition-Only Spiking Self-Attention (A\textsuperscript{2}OS\textsuperscript{2}A) framework effectively tackles the limited information capacity of binary spike activations through a mixed neuron strategy, another fundamental limitation persists in conventional spiking neural networks: inefficient quantization between weights and activations. Standard SNNs typically use binary spike activations combined with full-precision weights \cite{ReverB-SNN}. The underlying assumption is that binary spiking inherently provides superior energy efficiency by enabling addition-only operations, as each neuronal firing can replace costly multiply–accumulate operations with simple additions. However, this design disproportionately restricts representational capacity. Empirical findings in neural quantization research have demonstrated that reducing the precision of activations causes significantly more degradation in model accuracy than reducing the precision of weights\cite{ReverB-SNN}. Consequently, the conventional allocation of precision—favoring weights while retaining binary activations-may not offer the most effective trade-off between accuracy and efficiency. To overcome this structural constraint, ReverB-SNN \cite{ReverB-SNN} redefines quantization within SNNs by flipping the bit allocation between weights and activations. Instead of using binary spikes with full-precision synaptic weights, it introduces real-valued spike activations paired with binary weights. This reversal enhances the information flow within the network, as real-valued activations capture richer and more continuous variations in neuronal membrane potentials, allowing a more expressive and fine-grained representation of input signals. The neurons follow a modified Leaky-Integrate-and-Fire (LIF) mechanism, where the output activation corresponds to the membrane potential whenever the firing threshold is surpassed. Thus, each neuron conveys graded signal intensities rather than binary events, significantly boosting its feature encoding capability while preserving the asynchronous event-driven nature fundamental to SNNs. Despite employing real-valued output spikes, the network retains the critical multiplication-free property that defines energy-efficient spiking computation. This is achieved by constraining the weight parameters to binary values $\{-1, +1\}$. With this setup, synaptic operations reduce to simple addition or subtraction depending on the sign of the binary weight, fully replacing the need for floating-point multiplication during inference. This ensures that the efficiency advantages of SNNs remain intact. Moreover, to mitigate the inherent loss of flexibility from binary weights, ReverB-SNN integrates learnable scaling coefficients at the channel level during training \cite{ReverB-SNN}. These coefficients adaptively modulate the effective amplitude of each binary weight, enabling the learning process to approximate the continuous optimization dynamics of traditional networks without introducing substantial computational overhead. Once training is complete, these scaling parameters are merged into the corresponding activations through a re-parameterization procedure, restoring an inference model that remains entirely addition-based. 

Despite achieving fully spike-driven computation and maintaining energy efficiency, existing spiking Transformer architectures-whether employing matrix multiplications in Self-Attention \cite{Additon_only_Spiking_Transformer, Spikformer} or Hadamard products/masking operations \cite{QKFormer}-still rely on dot-product or element-wise similarity measures that were originally designed for dense, real-valued representations in ANNs \cite{Spiking_Vit_with_Saccadic_Attention}. The fundamental limitation arises from a mismatch between these operations and the intrinsic characteristics of binary spike trains\cite{Spiking_Vit_with_Saccadic_Attention}. Two critical problems emerge when query and key vectors are represented as binary spike sequences and the dot-product operation is used to measure similarity: First, binary spikes exhibit severe magnitude disparity and sparsity: each token vector contains only a small number of active spikes relative to the total dimensionality, resulting in highly variable spike counts across tokens \cite{Spiking_Vit_with_Saccadic_Attention}. This variability introduces significant randomness into similarity scores, making dot-product measurements unreliable indicators of semantic relevance. Second, when dot-product attention relies on angular similarity through cosine measures \( \cos \theta_{ij} = \frac{Q_i \cdot K_j^\top}{\| Q_i \| \| K_j \|} \), the small integer spike counts in the denominator cause extreme instability-minor timing variations or single spike differences can drastically alter the computed angle, even when the underlying semantic relationship remains constant. Masking operations further compound these issues by imposing hard thresholds on already noisy similarity estimates, distorting relevance assessments and preventing accurate spatio-temporal correlation capture.
To address this fundamental incompatibility between dot-product similarity and binary spike characteristics, the Spiking Vision Transformer with Saccadic Attention \cite{Spiking_Vit_with_Saccadic_Attention} replaces conventional relevance computation with a distribution-based cross-entropy approach tailored to spike trains. Spatially, the Saccadic Spike Self-Attention (SSSA) mechanism treats each token as a discrete firing distribution and computes relevance through cross-entropy between query and key firing rates as \( H(q, k) \approx - p_q \log p_k \). This formulation eliminates magnitude sensitivity by focusing solely on relative firing probabilities rather than vector norms, ensuring that similarity estimates remain robust despite sparse and variable spike counts. To avoid nonlinear logarithmic operations incompatible with spike-driven computation, the method further simplifies to \( \text{CroAtt}(Q, K) = - Q \bar{K}^\top \), where \( \bar{Q} \) and \( \bar{K} \) represent dimension-summed spike counts, achieving distribution-based relevance through efficient parallel addition-only operations \cite{Spiking_Vit_with_Saccadic_Attention}.

 Despite impressive progress in energy-efficient computation and architectural innovation, existing spiking Transformer models remain fundamentally limited by one crucial aspect: biological plausibility. As emphasized in the essay “Neuromorphic is Dead. Long Live Neuromorphic,” \cite{Neuromorphic_is_Dead} the term neuromorphic has drifted away from its original meaning. Instead of uncovering how brains compute, the field has become increasingly benchmark-driven, optimizing architectures that mimic AI rather than neurobiology \cite{Neuromorphic_is_Dead}. Current Spiking Transformers, for example, still depend on non-biological mechanisms such as addition-only matrix multiplication and cross-entropy-based computations-operations that have no direct counterpart in neural circuits. This drift risks losing what made neuromorphic engineering revolutionary in the first place: the belief that biological computation offers not just incremental improvements, but entirely new paradigms of learning, generalization, and adaptation. A return to these foundations calls for neuroAI, an interdisciplinary approach aimed at grounding learning rules and architectures in neuroscience rather than digital approximation \cite{Neuromorphic_is_Dead}. In contrast, the brain computes relevance between neural signals through the precise timing of spikes, not their magnitude or averaged firing rate. In summary, the existing spiking Transformers (i) still use ANN-style dot-product or cross-entropy attention mismatched to spike timing, (ii) rely on explicit attention score matrices that has limitations of von Neumann memory bottleneck, and (iii) remain weakly grounded in biological mechanisms. In this work, we revisit and try to solve the Transformer architecture through a neuromorphic lens, introducing a biologically plausible Spiking Transformer in which the attention mechanism emerges entirely from spike-timing-dependent plasticity (STDP)\textit{, }temporal coding, and simple additive operations-eliminating the need for softmax. The beauty of STDP is that it is the basis of memory, learning and intelligence in biological brain, and is widely studied in neuromorphic materials and devices. By grounding the attention computation in spike timing rather than magnitude, our framework moves beyond existing SNN-based Transformers and aligns more closely with the computational principles observed in real neural circuits. Our main contributions are as follows:
\begin{enumerate}
    \item This work replaces the conventional attention weight calculation between the query and key matrices with an attention score computed via spike-timing-dependent plasticity (STDP). This biologically plausible mechanism relies on local spike-timing interactions, encoding information salience directly in precise spike timing instead of magnitude. This enables efficient, event-driven processing precisely aligned with how brains compute synaptic relevance.
    
    \item The model leverages the spiking nature of the Value matrix to achieve inherently addition-only operations. By avoiding global operation such as softmax, the framework uses simple additive operations throughout the attention mechanism.

    \item By embedding the attention computation directly within synaptic plasticity dynamics, this approach eliminates the intermediate attention score matrix that must be explicitly materialized and stored in memory in prior spiking transformers. This circumvents the critical memory bottleneck encountered during the computation phase, substantially reducing memory bandwidth requirements and enabling efficient deployment on neuromorphic hardware with limited memory hierarchy.
    
\end{enumerate}
These prior works, together with our proposed approach, are summarized in Table~\ref{Tab1}. Neuromorphic computing was never meant to compete AI race \cite{Neuromorphic_is_Dead}. By returning to the principles that inspired the field, we aim to build architectures that not only perform well but also resonate with how the brain truly computes.

\begin{table*}[t]
\small
\centering
\caption{ Comparison of various spiking transformer architectures in terms of attention mechanisms, residual network, and biological plausibility.}
\label{Tab1}
\renewcommand{\arraystretch}{0.1} 
\begin{tabular}{p{2.8cm} p{1.7cm}  p{2.5cm}  p{1.5cm} c c c} 
\\[1pt]\hline\\[1pt]
\textbf{Method} & \textbf{\makecell{Attention\\Type}}  & \textbf{\makecell{Attention\\Module}} & \textbf{\makecell{Residual\\Conn. }} & \textbf{\makecell{Softmax}} & \textbf{\makecell{Brain\\inspired\\Attention }} \\ 
\\[1pt]\hline\\[1pt]
% --- Transformer(ANN) ---
\raggedright Transformer (ANN)\cite{r1} & \raggedright Spatial& VSA: $(QK^{T}\cdot V)s$ (Matrix Mul.)& Floating point & Yes  & No \\
\\[1pt]\hline\\[1pt]
% --- Spikeformer---
\raggedright Spikeformer\cite{Spikeformer} & \raggedright Spatial & SSA: $(QK^{T}\cdot V)s$ (Matrix Mul.) & Spike add (MultiBit)  & No  & No \\
\\[1pt]\hline\\[1pt]
% --- Spikformer ---
\raggedright Spikformer \cite{Spikformer} & \raggedright Spatial and \\Temporal  &SSA: $(QK^{T} \cdot V)s$ (Matrix Mul.) &Spike add (MultiBit) & No & No \\
\\[1pt]\hline\\[1pt]
% --- Spikingformer ---
\raggedright Spikingformer\cite{Spikingformer}  & \raggedright Spatial& SSA: $(QK^{T} \cdot V)s$ (Matrix Mul.) &Membrane potential (Binary)& No  & No \\
\\[1pt]\hline\\[1pt]
% --- TET ---
\raggedright QK-Former \cite{QKFormer} & \raggedright Spatial &Mask-operation (between Q,K) &Membrane potential (Binary) & No & No \\
\\[1pt]\hline\\[1pt]
% --- MS-ResNet --- (Multi-row entry)
\raggedright SpikingTransformer (A$^{2}$OS$^{2}$A)~\cite{Additon_only_Spiking_Transformer} & \raggedright Spatial & Hadamard Product (Binary Q, ReLU K, Ternany V)  & Membrane potential (Binary) & No & No \\
\\[1pt]\hline\\[1pt]
\raggedright Spiking ViT with Saccadic Attention \cite{Spiking_Vit_with_Saccadic_Attention} & \raggedright Spatial and Temporal& Matrix multiplication (Simplified Cross Entropy) & Membrane potential (Binary) & No  & No\\
\\[1pt]\hline\\[1pt] 
% --- $S^2TDPT$ --- (3 architectures)
\raggedright \textcolor{orange}{\textbf{Spiking STDP}\\\textbf{Transfromer (Ours)}} 
& \raggedright \textcolor{orange}{\textbf{Spatial}} 
& \raggedright \textcolor{orange}{\textbf{STDP kernel and first-spike time}}
& \textcolor{orange}{\textbf{Membrane potential (Binary)}} 
& \textcolor{orange}{\textbf{No}} 
& \textcolor{orange}{\textbf{Yes}} \\
\\[1pt]\hline\\[1pt] 
\end{tabular}
\end{table*}

\section{Methods}
\subsection{Spatial relevance Computation from spike Distribution}
Let $\mathbf{Q}$ and $\mathbf{K}$ be two $D$-dimensional binary vectors in the set $\{0,1\}^D$, representing two distinct spike trains or feature patterns:
\[
\mathbf{Q}, \mathbf{K} \in \{0, 1\}^D.
\]
The total number of ones in each vector is analogous to the total firing rate or spike count within an observation window. This can be computed as the $\ell_1$ norm of the vectors:
\[
P_Q = \sum_{i=1}^{D} q_i \quad \text{and} \quad P_K = \sum_{i=1}^{D} k_i,
\]
where $P_Q$ and $P_K$ denote the total number of spikes (or active features) in $\mathbf{Q}$ and $\mathbf{K}$, respectively. To translate the rate information into a temporal representation suitable for spike-timing-dependent plasticity (STDP), we employ a First-Spike Coding scheme \cite{s11,s12}. In this paradigm, a higher firing rate (or a larger $\ell_1$ norm) is encoded by an earlier spike time (shorter latency) relative to an arbitrary reference point $t_{\text{ref}}$. The mapping from the total spike count $P$ to spike time $t$ follows an inverse relationship. Assuming a linear or monotonic inverse relation, the spike times $t_Q$ and $t_K$ can be expressed as
\[
t_Q = \mathcal{F}(P_Q) \quad \text{and} \quad t_K = \mathcal{F}(P_K),
\]
where $\mathcal{F}:\mathbb{R}\rightarrow \mathbb{R}^+$ is a monotonically decreasing function such that larger $P$ yields smaller $t$ (earlier spike). The spike latency $t$ is now defined by a linear decay function based on the ratio of active features to the total dimensionality
\[
\mathcal{F}(P) = T_{\max} \cdot \left(1 - \frac{P}{D}\right)
\]
where $\mathbf{t}_Q$ is the latency of the $\mathbf{Q}$-encoded neuron (presynaptic), and $\mathbf{t}_K$ is the latency of the $\mathbf{K}$-encoded neuron (postsynaptic). Here, $T_{\max} $ denotes the maximum possible latency (which occurs when $P = 0$), and $D$ is the dimensionality of the vector, representing the maximum possible rate ($P_{\max} = D$). The mapping satisfies the constraint $0 \le t(P) \le T_{\max}$, where a greater $P$ yields a smaller (earlier) $t$. Consequently, if $\mathbf{Q}$ exhibits a higher firing rate than $\mathbf{K}$ ($P_Q > P_K$), its corresponding neuron (presynaptic) will fire earlier than that of $\mathbf{K}$ (postsynaptic), implying $t_Q < t_K$.
The temporal relation between these two spike events is quantified by the spike-time difference:
\[
\Delta t = t_Q - t_K.
\]
If $\Delta t < 0$ ($t_Q < t_K$), the presynaptic spike ($\mathbf{Q}$) precedes the postsynaptic spike ($\mathbf{K}$), resulting in Long-Term Potentiation (LTP). Conversely, if $\Delta t > 0$ ($t_Q > t_K$), the postsynaptic spike precedes the presynaptic spike, leading to Long-Term Depression (LTD)~\cite{r68}. The synaptic modification $\Delta w$, governed by the STDP mechanism, is used here as a similarity score $\mathcal{S}(\mathbf{Q}, \mathbf{K})$ between the two vectors, analogous to the dot product in continuous vector spaces. The standard asymmetric exponential STDP rule defines this similarity as
\[
\mathcal{S}(\mathbf{Q}, \mathbf{K}) \equiv \Delta w =
\begin{cases}
A_+ \exp{\!\left(\frac{\Delta t}{\tau_+}\right)}, & \text{if } \Delta t < 0 \quad (\text{Pre-before-Post: LTP}), \\[6pt]
- A_- \exp{\!\left(\frac{-\Delta t}{\tau_-}\right)}, & \text{if } \Delta t \ge 0 \quad (\text{Post-before-Pre: LTD}),
\end{cases}
\]
where $A_+$ and $A_-$ are positive constants representing the maximal amplitudes of potentiation and depression, respectively, and $\tau_+$ and $\tau_-$ are positive time constants determining the exponential decay of the respective effects.

\begin{figure}[htbp]
    \centering
    \includegraphics[width=0.9\textwidth]{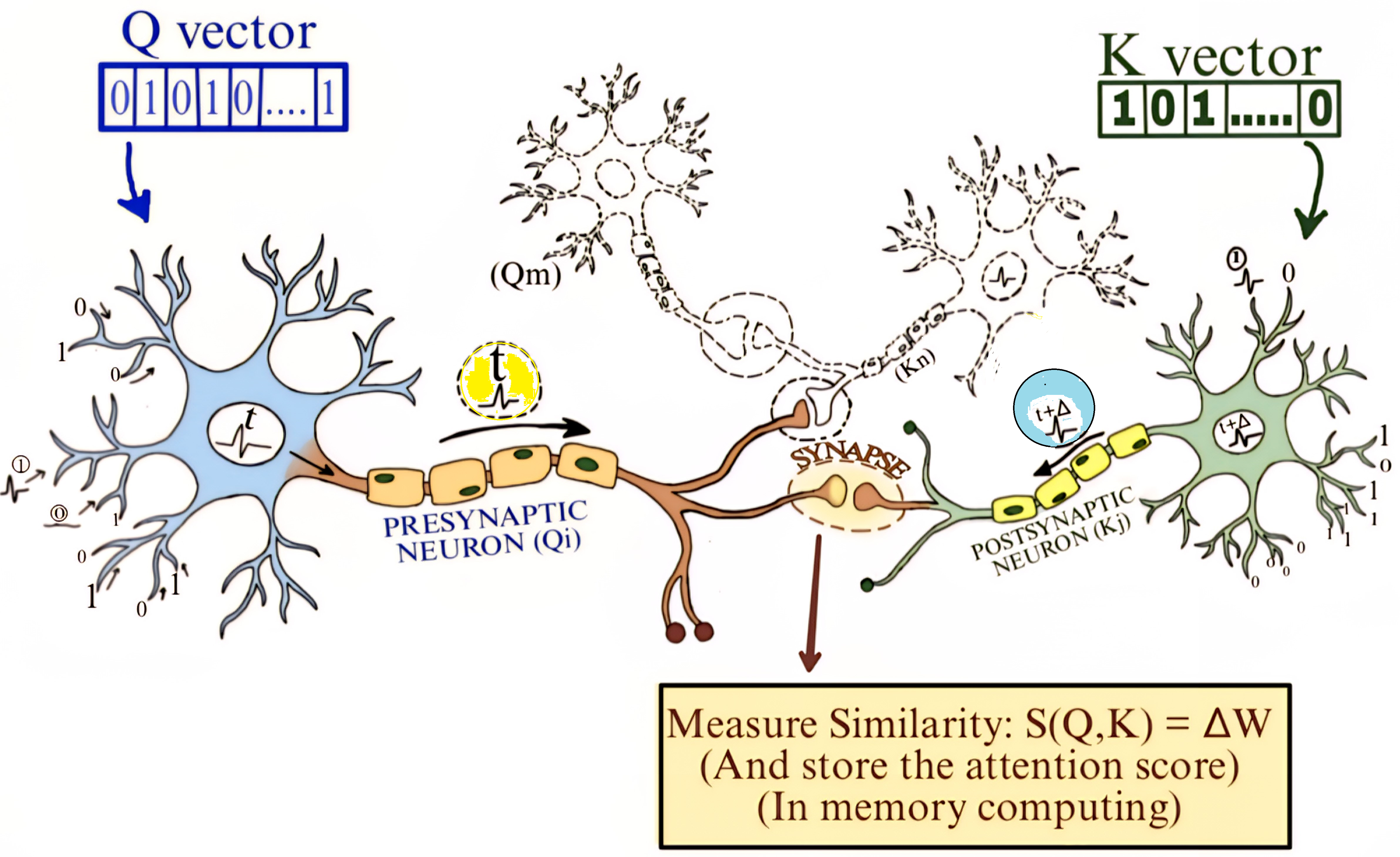}
    \caption{Synaptic similarity computation through spike-timing-dependent plasticity (STDP) and In-memory computing}
    \label{Fig1}
\end{figure}
\noindent The scalar $\Delta w$ thus quantifies the degree of similarity between $\mathbf{Q}$ and $\mathbf{K}$: a larger positive $\Delta w$ indicates stronger temporal correlation, reflecting that the presynaptic pattern $\mathbf{Q}$ reliably precedes (predicts) the postsynaptic pattern $\mathbf{K}$. 
% This provides a timing-sensitive, distribution-aware similarity measure that captures fine temporal dependencies between spike patterns, addressing the limitations of conventional spatial similarity measures when applied to sparsely distributed representations.
\noindent Unlike previous spiking transformer models, the proposed transformer does not maintain the similarity score $\Delta w$ as a value that is explicitly computed, stored, or accessed separately in memory during processing (Detailed information can be found in Appendix~\ref{A2})
Instead, this similarity measure is intrinsically embedded within the synapse through synaptic plasticity governed by the STDP mechanism. As a result, both the computation and storage of the $Q$-$K$ similarity are performed locally within the synaptic hardware. This co-localization of data processing and memory embodies the fundamental principle of in-memory computing, where each synapse simultaneously acts as both a computational and a storage element. The overall process of converting spike distributions into temporal latencies and subsequently computing synaptic similarity through STDP is illustrated in Fig.~\ref{Fig1}.

\subsection{Leaky Integrate-and-Fire Neuron Dynamics}
The spiking neuron layer serves as the core computational unit of the network, responsible for integrating information across both spatial and temporal domains. Incoming visual stimuli are accumulated in the membrane potential over time, and when the potential surpasses a firing threshold, discrete binary spikes are generated. This mechanism enables efficient temporal coding and event-driven computation. The behavior of spiking neurons \cite{maass1997}, including the evolution of their membrane potential and spike firing mechanism, is mathematically described in discrete time by the following recursive equations:
\begin{align}
U[t] &= H[t-1] + X[t], \\
S[t] &= \Theta(U[t] - V_{\text{th}}), \\
H[t] &= V_{\text{reset}} \cdot S[t] + \beta \cdot U[t] \cdot (1 - S[t]),
\end{align}
where $X[t]$ denotes the synaptic input current at timestep $t$, 
$U[t]$ represents the total membrane potential integrating spatial input with temporal history $H[t-1]$, 
$\Theta(\cdot)$ is the Heaviside step function producing a binary spike output $S[t]$ whenever the potential exceeds the firing threshold $V_{\text{th}}$. 
After firing, the membrane potential is reset to $V_{\text{reset}}$, while $\beta$ determines the rate at which the potential decays over time. For simplicity, the entire spiking neuron operation is represented as $SN(\cdot)$, which maps membrane potential tensors to spike tensors.

\subsection{Overall Architecture}

As depicted in Fig.~\ref{fig:spiking_transformer_overview}, the proposed brain-inspired Spiking Transformer is composed of four key components: Spiking Patch Splitting (SPS), Spiking STDP Self-Attention (S²TDPSA), a MultiLayer Perceptron (MLP), and a temporal aggregation classification head. The SPS module encodes visual information into spike-based representations, enabling efficient extraction of spatial and temporal features. The S\textsuperscript{2}TDPSA module integrates a Spike Time Dependent Plasticity (STDP)-based attention mechanism that captures long-range dependencies, where different elements of \( V \) are dynamically weighted according to spike-timing correlations between \( Q \) and \( K \), resulting in biologically plausible attention computation. The MLP block further refines these encoded features through spike-based feed-forward processing, enhancing the network’s representational capacity. Finally, the classification head aggregates spiking activity across timesteps to produce the final decision output. Each of these components is described in detail in the subsequent sections. The proposed architecture builds upon the framework introduced in Spikingformer \cite{Spikingformer}. The network processes 2D image sequences organized as a spatiotemporal tensor $I \in \mathbb{R}^{T \times C \times H \times W}$, where $T$ denotes the number of simulation timesteps over which information propagates through recurrent membrane dynamics, $C$ represents the number of input channels, and $H$ and $W$ specify the height and width of the image sequence. Our design philosophy integrates recent developments in spiking transformer architectures with a novel plasticity-driven attention mechanism, 
facilitating fully event-driven and addition-only computation that is well suited for neuromorphic hardware implementation.

\begin{figure}[htbp]
    \centering
    \includegraphics[width=0.9\textwidth]{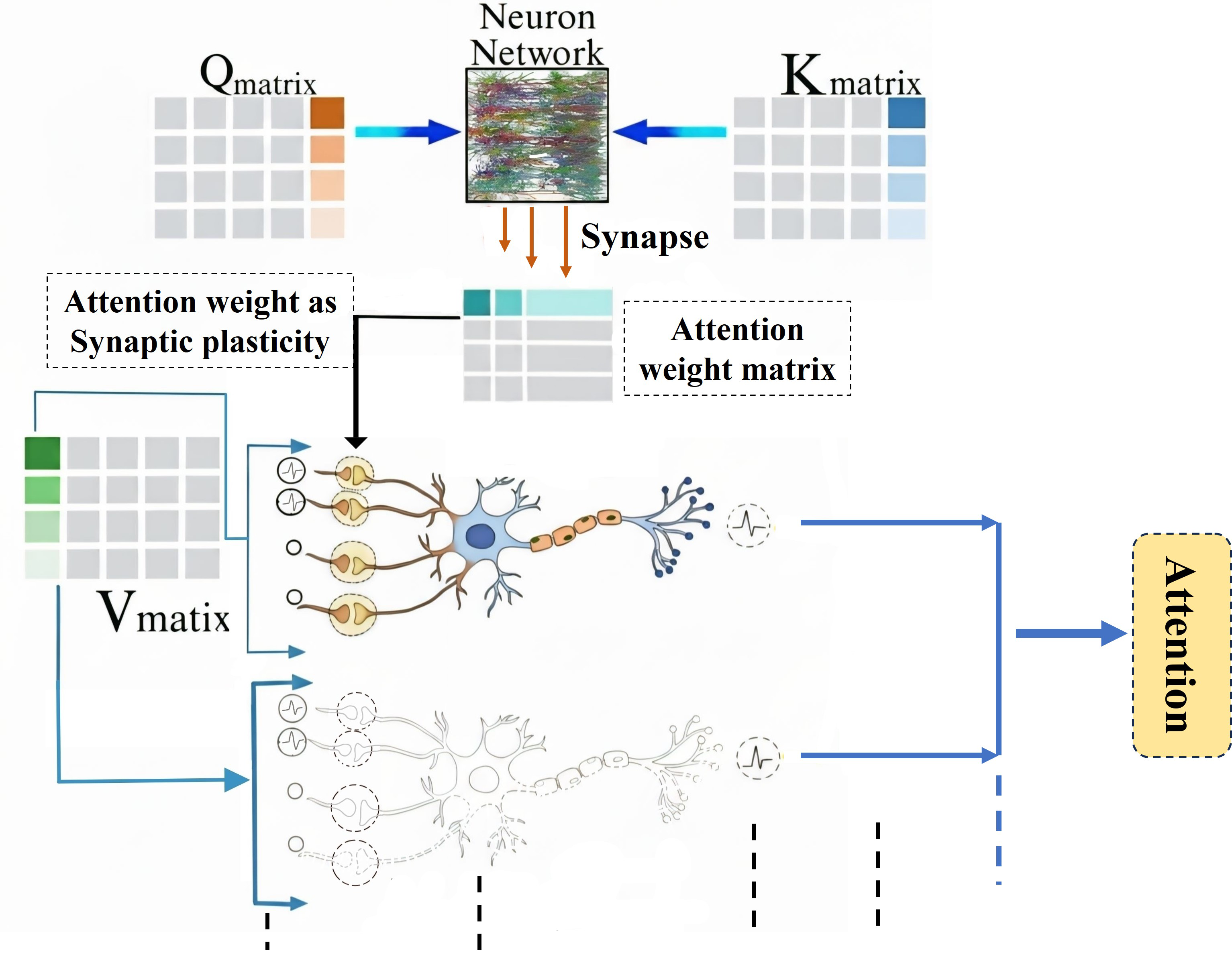}
    \caption{Brain-inspired attention module proposed in this work. The neural network computes the attention weight matrix by capturing similarity between the Q and K matrices using STDP, which is then used to weight the V matrix according to importance.}
    \label{FIG2}
\end{figure}

\begin{figure}[htbp]
    \centering
    \includegraphics[width=0.9\textwidth]{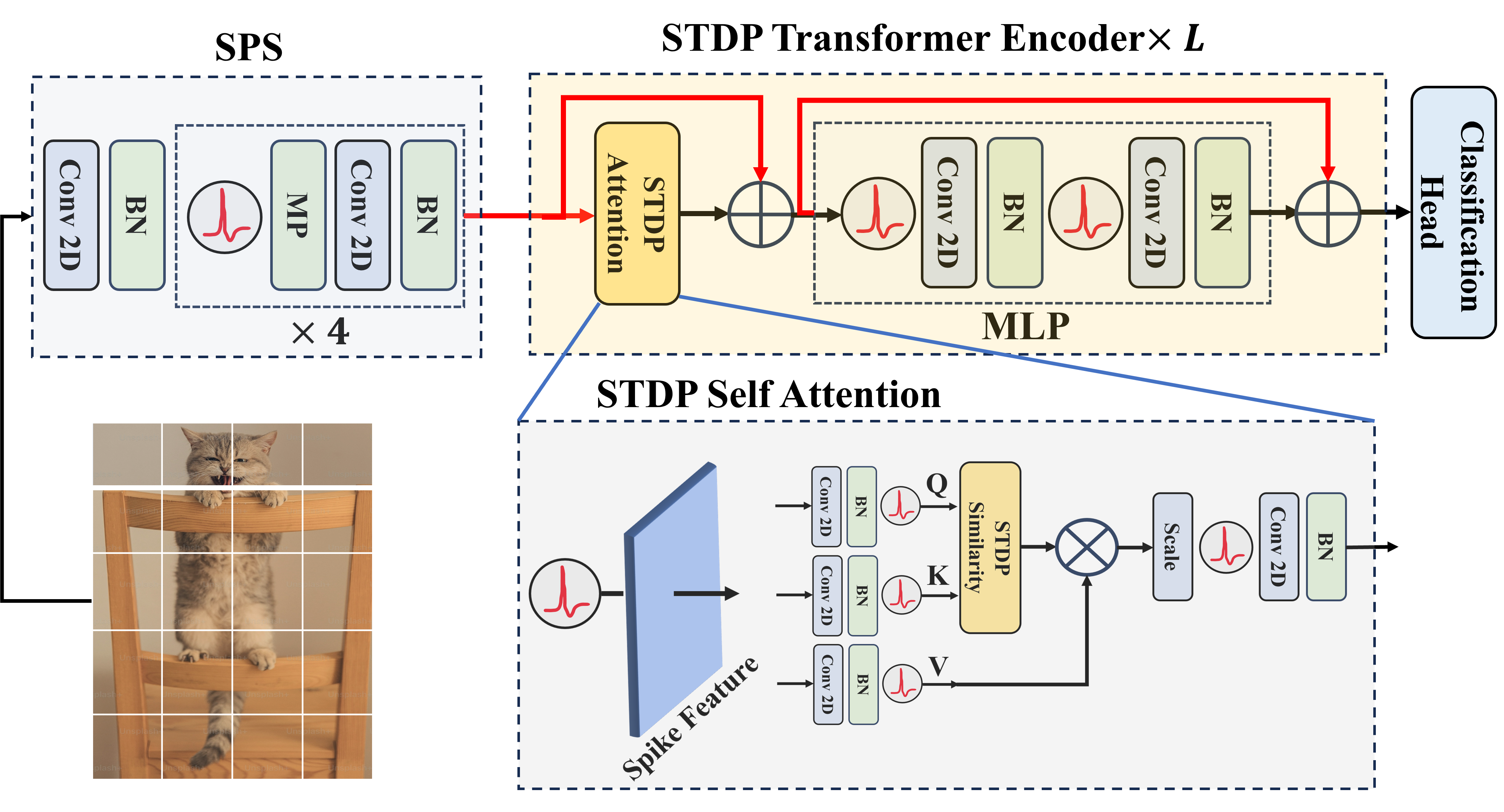}
    \caption{Overview of the Spiking STDP Transformer, comprising Spiking Patch Splitting (SPS), an encoder layer with STDP-based self-attention and a multi-layer perceptron (MLP) module, followed by the classification head. The STDP Self Attention module is proposed in this work.}
    \label{fig:spiking_transformer_overview}
\end{figure}

The Spiking Patch Splitting (SPS) module is designed to convert raw visual inputs into spike-based patch tokens by feature extraction while simultaneously performing patch embedding and spatial compression. Given an input 
\(I \in \mathbb{R}^{T \times C \times H \times W}\), the module generates a progressively refined feature representation through consecutive convolutional layers\cite{Spikingformer}. The Spiking Patch Splitting integrates a convolutional spiking embedding stage with a downsampling mechanism that reduces the feature map to a fixed resolution. When static images are provided, the initial convolution block acts as a spike encoder, converting pixel intensities into temporal spike trains. A standard 2D convolution with $3 \times 3$ kernal and stride~1 is denoted as $\mathrm{Conv2d}$, while batch normalization, max-pooling (stride~2), and the multi-step spiking neuron model are represented as $\mathrm{BN}$, $\mathrm{MP}$, and $\mathrm{SN}$, respectively. Equations~(11) and~(12) describe the two forms of spiking patch embedding employed in the tokenizer: the first maintains the original spatial resolution (SPE), while the second integrates a downsampling operation (SPED). In the SPS module, multiple SPE and SPED layers are arranged hierarchically to accommodate classification tasks that require varying levels of spatial downsampling.
The overall SPS operation can be formalized as follows:
\begin{center}
\begin{align}
\mathbf{Z}_0 &= \mathrm{BN}\!\left(\mathrm{Conv2d}(\mathbf{I})\right), \\[6pt]
\mathbf{Z}_{\mathrm{SPE}} &= \mathrm{BN}\!\left(\mathrm{Conv2d}\!\left(\mathrm{SN}(\mathbf{Z}_i)\right)\right), \\[6pt]
\mathbf{Z}_{\mathrm{SPED}} &= \mathrm{BN}\!\left(\mathrm{Conv2d}\!\left(\mathrm{MP}\!\left(\mathrm{SN}(\mathbf{Z}_j)\right)\right)\right), \\ 
\mathbf{U}_0 &= \mathrm{SPS}(\mathbf{I}), \quad 
\mathbf{I} \in \mathbb{R}^{T \times C \times H \times W}, \quad
\mathbf{U}_0 \in \mathbb{R}^{T \times N \times D}.
\end{align}
\end{center}

\noindent Here, $T$ denotes the number of simulation steps, $B$ the batch size, and $D$ the embedding dimension.
In the above equations, $\mathbf{U}_0$ represents the membrane potential obtained from the SPS, \(N\) denotes the total number of patches. This design ensures that all computations within the SPS module are implemented through spike-driven operations. The output $\mathbf{U}_0$ is subsequently forwarded to an $L$-layer Spiking Transformer encoder, where each encoder block is composed of two primary components: S²TDPSA module and a feed-forward multilayer perceptron (MLP). To prevent multi-bit spike generation, we follow the design principles outlined in \cite{Spikingformer,Spike-driven_Transformer_V2}, which constrain the pre-convolutional activations to binary values. This formulation ensures that spike-weight interactions are implemented through simple additive operations, thereby reducing the computational complexity of the convolutional process. All modules are connected via membrane potential residual links, which facilitate gradient propagation across depth while maintaining spike-based inter-layer information flow. The attention computation begins by projecting the normalized input spike sequence through three parallel branches to obtain the spiking query(${Q}_S$), key(${K}_S$), and value(${V}_S$) representations. Each projection branch comprises a linear transformation, batch normalization, and multi-step LIF spike generation, 
which can be expressed as:
\begin{align}
\text{Q}_S &= SN\big(BN(W_Q \tilde{\mathbf{S}}^{\ell-1})\big), \\
\text{K}_S &= SN\big(BN(W_K \tilde{\mathbf{S}}^{\ell-1})\big), \\
\text{V}_S &= SN\big(BN(W_V \tilde{\mathbf{S}}^{\ell-1})\big),
\end{align}
where $\mathbf{Q}_S, \mathbf{K}_S, \mathbf{V}_S \in \{0,1\}^{B \times N \times D}$, with $B$ representing the batch size.
The spike tensors are subsequently reshaped into a multi-head configuration $\mathbb{R}^{B \times H \times N \times D_H}$, where $H$ denotes the number of attention heads and $D_H = D / H$ specifies the dimensionality per head. This multi-head representation allows parallel spike-driven attention processing. For each token index $i$ and $j$, the spike rates of the query and key sequences are computed by summing the spike activations across feature dimensions:
\begin{align}
r_{Q}(i) &= \sum_{d=1}^{D_H} \mathbf{Q}_s(i,d), &
r_{K}(j) &= \sum_{d=1}^{D_H} \mathbf{K}_s(j,d),
\end{align}
where $(i, j)$ denote token indices and $d$ indexes the feature dimension within each attention head. To map spike rates to temporal domain representations, an inverse relationship is employed, where higher spike rates correspond to earlier firing times:
\begin{align}
t_{Q}(i) &= T_{\max}\left(1 - \frac{r_{Q}(i)}{D_H}\right), &
t_{K}(j) &= T_{\max}\left(1 - \frac{r_{K}(j)}{D_H}\right),
\end{align}
Temporal causality between all token pairs is captured through a timing difference matrix:
\begin{align}
\Delta t_{ij} = t_{Q}(i) - t_{K}(j), \quad \Delta t \in \mathbb{R}^{N \times N},
\end{align}
where positive $\Delta t_{ij}$ indicates that the query spike occurs after the key spike (causal), while negative values correspond to anti-causal relationships. The spike-timing dependent plasticity (STDP) kernel is then defined as:
\begin{align}
f_{\text{STDP}}(\Delta t) = A_{\text{stdp}} \exp\left(-\frac{|\Delta t|}{\tau_{\text{stdp}}}\right),
\end{align}
where $A_{\text{stdp}}$ controls the update amplitude and $\tau_{\text{stdp}}$ determines the temporal correlation window. 
The synaptic weight update rule is asymmetrical, reflecting biological long-term potentiation (LTP) and long-term depression (LTD):
\begin{align}
\Delta w_{ij} =
\begin{cases}
f_{\text{STDP}}(\Delta t_{ij}), & \text{if } \Delta t_{ij} <0 \quad \text{(anti-causal)} \\
- f_{\text{STDP}}(\Delta t_{ij}), & \text{if } \Delta t_{ij} \ge 0 \quad \text{(causal)}.
\end{cases}
\end{align}
This mechanism strengthens synaptic connections for anti-causal spike pairs ($\Delta t < 0$) via LTP and weakens them for causal pairs ($\Delta t \ge 0$) via LTD. This spike-timing asymmetry enables the model to calculate temporal similarity in an biologically plausible manner. To ensure that the attention score ($A_t$) remain positive, the synaptic updates $\Delta w_{ij}$ are shifted by a constant offset($w_{offset}$), which moves all values into a positive range:
\begin{align}
\text{A}_{ij} = \Delta w_{ij} + w_{\text{offset}},\text{$A_t$} \in \mathbb{R}^{B \times N \times N},
\end{align}
By properly tuning $A_{\text{stdp}}$ and $w_{\text{offset}}$, the attention scores are inherently bounded. The maximum value of $A_{\text{stdp}} \exp\left(-\frac{|\Delta t|}{\tau_{\text{stdp}}}\right)$ is $A_{\text{stdp}}$, and with an appropriate choice of $w_{\text{offset}}$, the attention values remain within the range $(0,1)$. This eliminates the need for a Softmax function commonly used in conventional attention mechanisms, as all attention weights are already positive \cite{Additon_only_Spiking_Transformer}. Since $V$ consists of spike-based values, i.e., binary entries $\{0,1\}$, and the attention weights $A_t$ are bounded between $(0,1)$, the resulting product remains naturally constrained \cite{Additon_only_Spiking_Transformer}. Consequently, the matrix multiplication effectively reduces to an addition-only operation in the spiking domain, preserving energy-efficient computation without the need for extra normalization \cite{Additon_only_Spiking_Transformer}. The attention mechanism proposed in this work is illustrated in Fig.~\ref{FIG2}. The output of the $S^2\text{TDPSA}$, is computed as :
\begin{align}
S^2\text{TDPSA}(\mathbf{Q}_s, \mathbf{K}_s, \mathbf{V}_s) 
= BN\big(\text{Conv2d}(SN(\mathbf{A}_t \cdot \mathbf{V}_s \cdot s))\big)\,
\end{align}
For the $l$-th Transformer encoder layer ($l = 1 \dots L$), the input is the membrane potential $\mathbf{U}_{l-1} \in \mathbb{R}^{T \times N \times D}$. We adopt $s$ as the scaling factor, analogous to the approach described in~\cite{r1}.The attention sub-block($S^2\text{TDPSA}$) computes the spike output as:
\begin{equation}
\mathbf{S}'_l = S^2\text{TDPSA}(\mathbf{Q}_s, \mathbf{K}_s, \mathbf{V}_s), \mathbf{S}'_l \in \mathbb{R}^{T \times N \times D},
\end{equation}

\noindent This output is integrated into the block's membrane potential via the first residual connection:
\begin{equation}
\mathbf{U}'_l = \mathbf{U}_{l-1} + \mathbf{S}'_l , \mathbf{U}'_l \in \mathbb{R}^{T \times N \times D}.
\end{equation}
The updated membrane potential $\mathbf{U}'_l$ is normalized using LayerNorm and passed through the two-layer spiking MLP module. The MLP produces a output membrane potential $\mathbf{S}_l$, where $\mathbf{S}_{\text{hid}}$ is the hidden layer's response:
\begin{equation}
\mathbf{S}_{l} =  \text{BN}_{\text{out}} \big( \text{Conv2d}_{\text{out}} (\text{SN}_{\text{out}}(\mathbf{S}_{\text{hid}}))\big)  ,  \mathbf{S}_{l} \in \mathbb{R}^{T \times N \times D}.
\end{equation}
The final membrane potential from MLP is obtained through a second residual link:
\begin{equation}
\mathbf{X}_{l} = \mathbf{U}'_l + \mathbf{S}_{l},  \mathbf{X}_{l} \in \mathbb{R}^{T \times N \times D}
\end{equation}
After propagating through all $L$ encoder layers, the last membrane potential $\mathbf{X}_l$ is converted into the classification output $\mathbf{Y}$. This involves \textbf{Global Temporal Mean Pooling (GTMP)} across the $T$ timesteps, followed by \textbf{Global Average Pooling (GAP)} across tokens, and projection through a fully connected \textbf{Classification Head (CH)}. Detailed information about GTMP and GAP can be found in Appendix \ref{A1}.
\begin{equation}
\mathbf{x}_{\text{final}} = \text{GAP} \Bigg( \frac{1}{T} \sum_{t=1}^{T} \mathbf{U}_{L}(t, \cdot, \cdot) \Bigg), \quad
\mathbf{Y} = \text{CH}(\mathbf{x}_{\text{final}}).
\end{equation}

\subsection{Theoretical analysis of Energy Consumption}

Due to the inherent property of convolutional layers, subsequent batch normalization (BN) and linear scaling operations can be mathematically merged with the convolution, resulting in a single convolutional layer that includes an additional bias term during deployment~\cite{Spikingformer,s11,s3,s40,s41}. As a result, batch normalization layers do not contribute to the overall theoretical energy consumption and can be omitted from such calculations. For the $S^2TDPT$ architecture, the total count of synaptic operations is determined prior to estimating energy consumption. Specifically, the number of spike-based operations at each block or layer $l$, denoted as $SOP_l$, is calculated as $SOP_l = f_l \cdot T \cdot FLOPs_l$, where $f_l$ represents the firing rate, $T$ is the simulation time window, and $FLOPs_l$ is the number of floating-point operations (principally multiply-and-accumulate, or MAC, operations) associated with that block or layer \cite{Spikingformer}. Here, $SOP_l$ captures the total spike-driven accumulate (AC) computations.

The theoretical energy requirements for $S^2TDPT$ are estimated based on methodologies described in prior works~\cite{B1,B2,B3,B4,B5,B6,B7}. Calculations assume 45nm CMOS hardware, with the energy per MAC ($E_{MAC}$) set at 46~pJ and per AC ($E_{AC}$) at 0.9~pJ, following the conventions of~\cite{B2}. The overall theoretical energy consumption for $S^2TDPT$, when processing static datasets with RGB image input, is given by: 

\begin{equation}
E_{rgb}^{S^2TDPT} = E_{AC} \sum_{i=2}^{N} SOP_i^{Conv} + \sum_{j=1}^{M} SOP_j^{SSA} + E_{MAC} \cdot FLOP_1^{Conv}
\label{eq:energy_consumption}
\end{equation}
Equation~\eqref{eq:energy_consumption} describes the total energy cost, where $FLOP_1^{Conv}$ refers to the conversion of the initial RGB image into spike representation by the first SPS layer. The subsequent energy is computed by summing all synaptic operations in both convolutional and self-attention layers-each multiplied by their respective per-operation energy cost. BN energy is excluded as explained above, and all operations are summed to provide a comprehensive and hardware-relevant estimation of energy usage in $S^2TDPT$.

\section{Results}

\subsection{CIFAR Classification}
The \textbf{CIFAR-10 and CIFAR-100}\cite{cifar} datasets each consist of 50,000 training and 10,000 testing images, with a spatial resolution of $32 \times 32$ pixels. The key distinction lies in the number of classification categories: CIFAR-10 comprises 10 classes, whereas CIFAR-100 includes 100 fine-grained categories, offering a more challenging benchmark and higher discriminative capacity for evaluating classification algorithms. In our experimental setup, the batch size for \textbf{$S^2TDPT$} is fixed at 64. Within the SPS module, two Spiking Patch Embedding Stages (SPEs) and two Spiking Patch Expansion and Downsampling (SPED) units are employed to divide the input images into 64 non-overlapping patches of size $4 \times 4$. The corresponding experimental outcomes are summarized in Table~\ref{Tab2}.

\begin{table}[h]
\normalsize % Keeps font small, but can be changed to \normalsize
\centering
\caption{Comparison of performance on CIFAR-10 and CIFAR-100 datasets. ``Param'' denotes the number of learnable parameters (in millions), and ``Time Step'' represents the number of discrete simulation timesteps used during inference. Here, Architecture-$L$--$D$ denotes Transformer architectures with $L$ Transformer encoder blocks and $D$-dimensional feature embeddings.
}
\label{Tab2}
\renewcommand{\arraystretch}{0.3} % Reduce default vertical padding for minimal gap
% Column specifiers: Method (2.5cm), Architecture (2.5cm), Params (c), TimeStep (c), CIFAR10 (c), CIFAR100 (c)
\begin{tabular}{p{3.7cm} p{2.9cm}  c c c c c} 
\\[1pt]\hline\\[1pt]
\textbf{Method} & \textbf{Architecture} & \textbf{\makecell{Param\\(M)}} & \textbf{\makecell{Time\\Step}} & \textbf{\makecell{CIFAR10\\Top-1Acc(\%)}} & \textbf{\makecell{CIFAR100\\Top-1Acc(\%)}} \\ 
\\[1pt]\hline\\[1pt]
% --- Hybridtraining ---
\raggedright Hybridtraining(2018)~\cite{r53} & \raggedright VGG-11 & 9.27 & 125 & 92.22 & 67.87 \\
\\[1pt]\hline\\[1pt]
% --- STBP ---
\raggedright STBP(2019)~\cite{r55} & \raggedright CIFARNet & 17.54 & 12 & 89.83 & -- \\
\\[1pt]\hline\\[1pt]
% --- STBPNeuNorm ---
\raggedright STBPNeuNorm(2020)~\cite{r56} & \raggedright CIFARNet & 17.54 & 12 & 90.53 & -- \\
\\[1pt]\hline\\[1pt]
% --- TSSL-BP ---
\raggedright TSSL-BP(2021)~\cite{r57} & \raggedright CIFARNet & 17.54 & 5 & 91.41 & -- \\
\\[1pt]\hline\\[1pt]
% --- DSpike ---
\raggedright Dspike(2021)~\cite{C4} & \raggedright ResNet-18 & -- & 6 & 94.30 & 74.20 \\
\\[1pt]\hline\\[1pt]
% --- tdBN ---
\raggedright tdBN(2020)~\cite{C1} & \raggedright ResNet-19 & 12.63 & 6/4 & 93.20 & 70.86 \\
\\[1pt]\hline\\[1pt]
% --- Diet-SNN ---
\raggedright DIET-SNN(2021)~\cite{C3} & \raggedright ResNet-20 & -- & 5 & 92.70 & 69.67 \\
\\[1pt]\hline\\[1pt]
% --- MS-ResNet --- (Multi-row entry)
\raggedright MS-ResNet(2021)~\cite{r7} & \raggedright ResNet-110 & -- & -- & 91.72 & 66.83 \\
& \raggedright ResNet-48 & -- & -- & 91.90 & -- \\
\\[1pt]\hline\\[1pt]
% --- SEW-ResNet ---
\raggedright SEW-ResNet(2009)~\cite{C2} & \raggedright SEW-ResNet-21/18 & -- & 16/6 & 90.02 & 77.37 \\
\\[1pt]\hline\\[1pt]
\raggedright FA-Bi-I(2022) ~\cite{y1} & \raggedright CNN & -- & --& 87 & -- \\
\\[1pt]\hline\\[1pt]
\raggedright $\mathrm{ReS}_2$(2021)~\cite{y2} & \raggedright CNN & -- & --& 89 & -- \\
\\[1pt]\hline\\[1pt]

\raggedright ANN(2017)~\cite{r1} & \raggedright ResNet-19 & 12.63 & - & 94.97 & 75.35 \\
& \raggedright Transformer-4-384 & 9.32 & - & 96.73 & 81.02 \\
\\[1pt]\hline\\[1pt]

\raggedright Spikformer\cite{Spikeformer}(2022) & \raggedright Spikformer-4-384 & 9.32 & 4 & 95.19 & 77.86 \\
\\[1pt]\hline\\[1pt]
\raggedright Spikingformer\cite{Spikingformer}(2023) &\raggedright Spikingformer-4-384 & 9.32 & 4 & 95.61 & 79.09\\

\\[1pt]\hline\\[1pt]
\raggedright Spiking-Transformer(2025)~\cite{Additon_only_Spiking_Transformer}
& \raggedright Spiking Transformer-4-384 & 9.32 & 4 & 96.32 & 79.69 \\
\\[1pt]\hline\\[1pt] 
\raggedright \textcolor{brown}{\textbf{Spiking STDP Transformer}} & 
\raggedright \textcolor{brown}{\textbf{$S^2TDPT$-4-384}} & 
\textcolor{brown}{\textbf{9.32}} & 
\textcolor{brown}{\textbf{4}} & 
\textcolor{brown}{\textbf{94.35}} & 
\textcolor{brown}{\textbf{78.08}} \\
\raggedright \textcolor{brown}{\textbf{($S^2TDPT$) (Ours)}} &
 & & & & \\
\\[1pt]\hline\\[1pt] 
\end{tabular}
\end{table}

Figure~\ref{F4} presents the interpretability analysis of the Spiking Transformer model using Spiking Grad-CAM \cite{r14} ($S^2$TDPT) and the Spike Firing Rate (SFR) map \cite{r67} for five representative samples (car, horse, dog, deer). The visualizations consistently indicate that the network performs localized, object-centered feature extraction. In all cases, the Grad-CAM results highlight activation regions that closely align with the object boundaries, confirming that the model’s classification decisions-derived from the final Transformer block’s feature maps-are guided primarily by the most discriminative areas of the input image. Furthermore, the SFR map provides a quantitative view of the model’s internal dynamics by depicting the average spike firing rate across all attention heads, time steps ($T=4$), and Transformer blocks (Depth:0-3). The resulting maps reveal that pixels corresponding to object regions exhibit the highest cumulative spiking activity (approaching~1.0), which shows strong spatial correspondence with the Grad-CAM saliency. This close alignment between the model’s decision focus and its internal spiking activity underscores the architecture’s capacity to extract compact, energy-efficient, and interpretable representations for reliable image classification.

 \begin{figure}[htbp]
    \centering
    \includegraphics[width=0.6\textwidth]{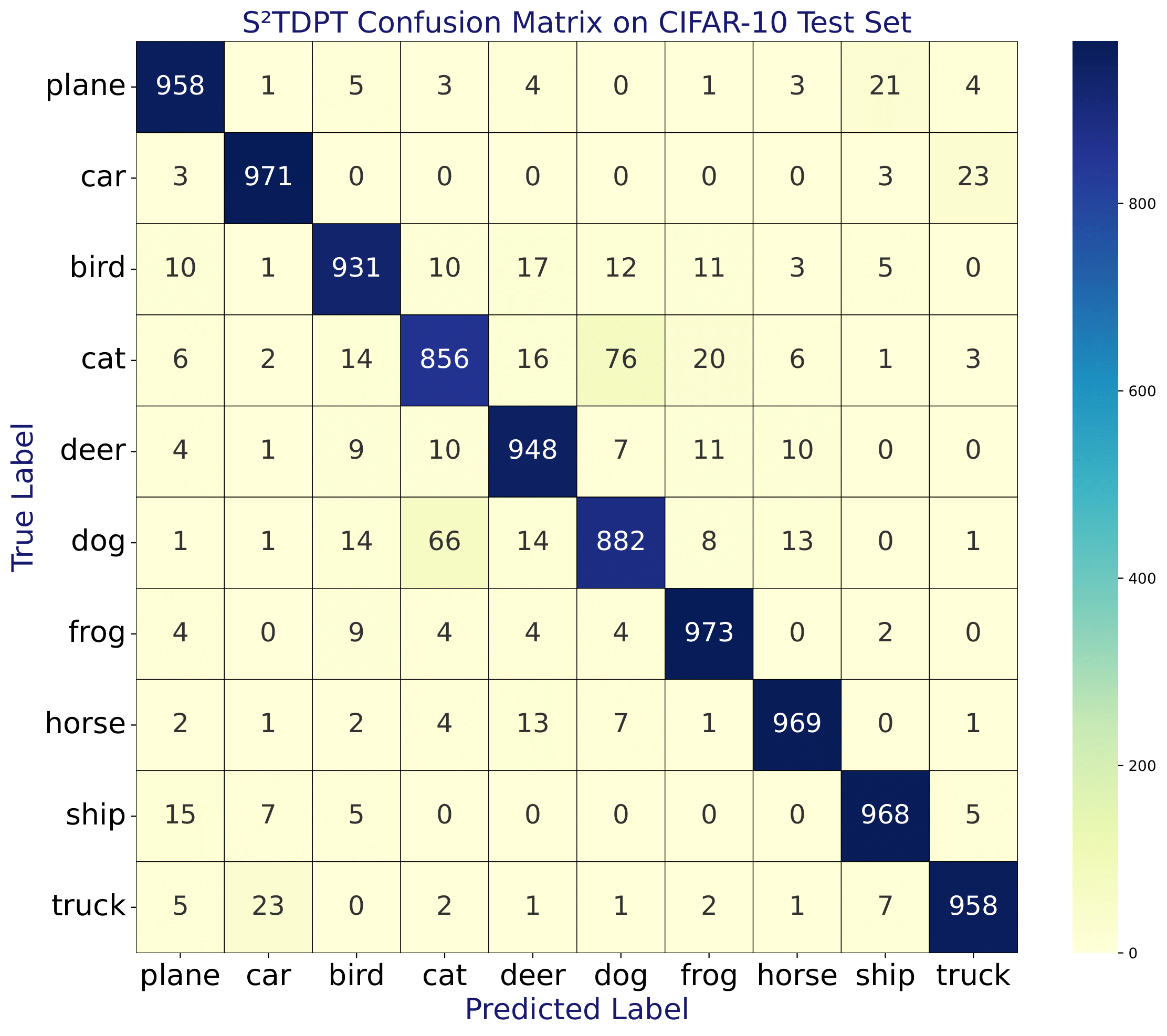}
    \caption{S\textsuperscript{2}TDPT confusion matrix on CIFAR-10, showing per-class classification performance.}
    \label{Fig4}
\end{figure}

When evaluated on the CIFAR-100 and CIFAR-10 datasets, $S^2TDPT$ achieved superior classification accuracy as shown in the Table~\ref{Tab2} while maintaining low energy consumption. With an energy consumption of 0.49~mJ, computed using Eq.~(\ref{eq:energy_consumption}), the model achieved top-1 accuracy of 78.08\% on CIFAR-100 and a top-1 accuracy of 94.35\% on CIFAR-10, surpassing multiple state-of-the-art methods.
 Specifically, $S^2TDPT$-4-384 achieves a top-1 classification accuracy of 78.08\% on the CIFAR-100 dataset using only four time steps, significantly outperforming Dspike~\cite{C4} by 3.88\%, ANN-ResNet19 \cite{C5} by 2.73\%, Spikformer-4-384 \cite{Spikformer} by 0.22\%, the MS-ResNet-110 model \cite{r7} by 11.25\%, and the SEW-ResNet\cite{C2} model by 0.71\%.

\begin{table*}[t]
\normalsize 
\centering
\caption{Comparison of Energy Consumption for the Transformer-4-384 Architecture for CIFAR-100 dataset}
\label{T4}
\renewcommand{\arraystretch}{0.3} 
\begin{tabular}{p{4cm}  c c} 
\\[1pt]\hline\\[1pt]
\textbf{Method} & \textbf{\makecell{Param\\(M)}} & \textbf{\makecell{Energy\\(mJ)}} \\ 
\\[1pt]\hline\\[1pt]
% --- Hybridtraining ---
\raggedright ANN Transformer~\cite{r1,01}  & 9.32& 4.25  \\
\\[1pt]\hline\\[1pt]
% --- STBP ---
\raggedright Spikformer~\cite{Spikformer,01} & 9.32 & 0.79 \\
\\[1pt]\hline\\[1pt]
% --- STBPNeuNorm ---
\raggedright SAFormer~\cite{SAFormer} & -- & 0.58 \\
\\[1pt]\hline\\[1pt]
% --- TSSL-BP ---
\raggedright S-Transformer~\cite{spikedriventransformer,SAFormer}& 9.32 & 0.6177 \\
\\[1pt]\hline\\[1pt]
\raggedright \textcolor{brown}{\textbf{Spiking STDP}} \\
\textcolor{brown}{\textbf{Transfromer}}  & \textcolor{brown}{\textbf{9.32}} & \textcolor{brown}{\textbf{0.49}} \\
\textcolor{brown}{\textbf{(}\textbf{$S^2TDPT$)(Ours)}} \\
\\[1pt]\hline\\[1pt] 
\end{tabular}
\end{table*}

\noindent The confusion matrix of S\textsuperscript{2}TDPT on CIFAR-10, shown in Fig.~\ref{Fig4}, provides a detailed view of per-class performance across the ten categories. These findings highlight the model’s effectiveness in handling complex, multi-category image classification tasks.

\begin{figure}[H]
    \centering
    \begin{minipage}{0.56\textwidth}
        \centering
        % Row 1: sample 1 (three visualizations side by side)
        \includegraphics[width=1.0\textwidth]{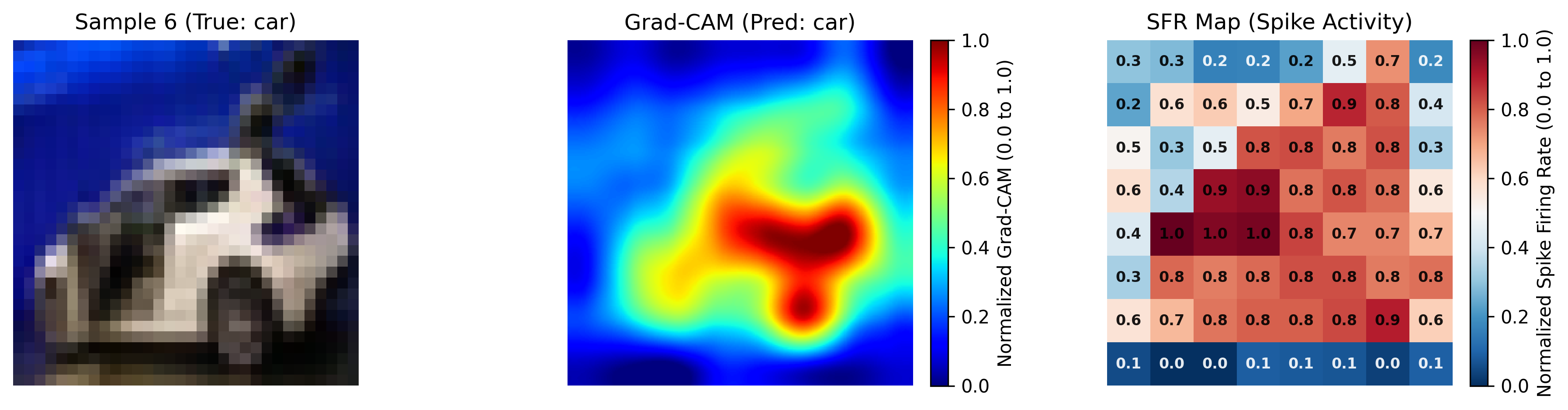}
       
        \vspace{1mm}

        % Row 2: sample 2
        \includegraphics[width=1.0\textwidth]{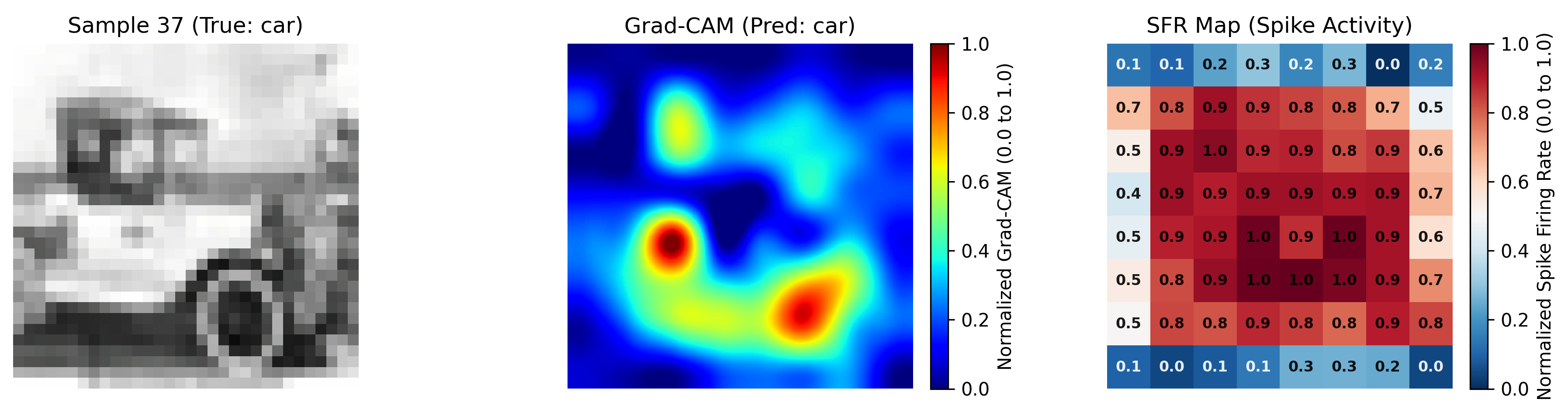}
       
        \vspace{1mm}

        % Row 3: sample 3
        \includegraphics[width=1.0\textwidth]{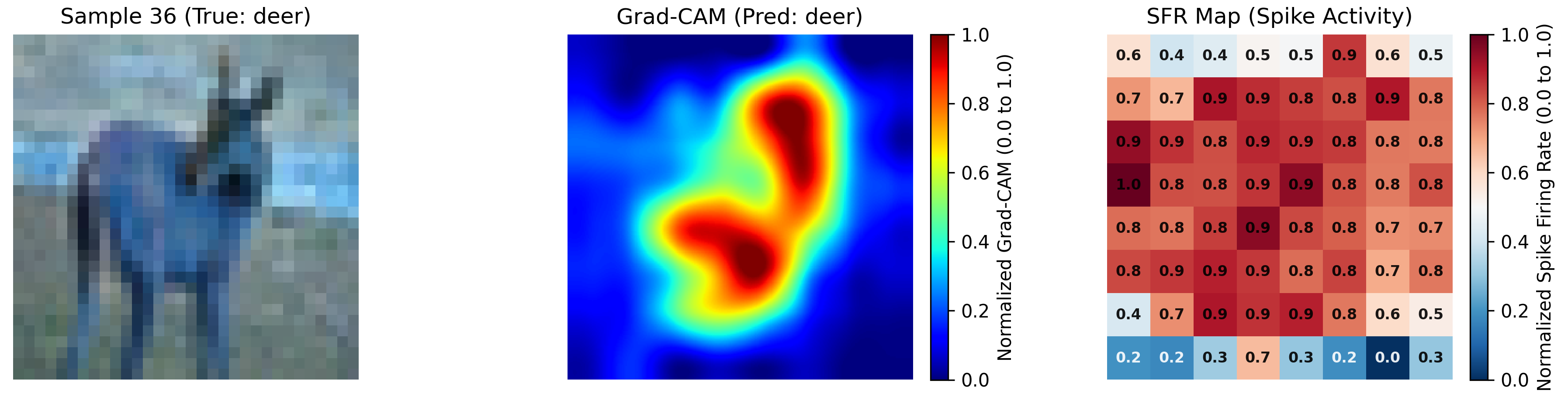}
       
        \vspace{1mm}

        % Row 4: sample 4
        \includegraphics[width=1.0\textwidth]{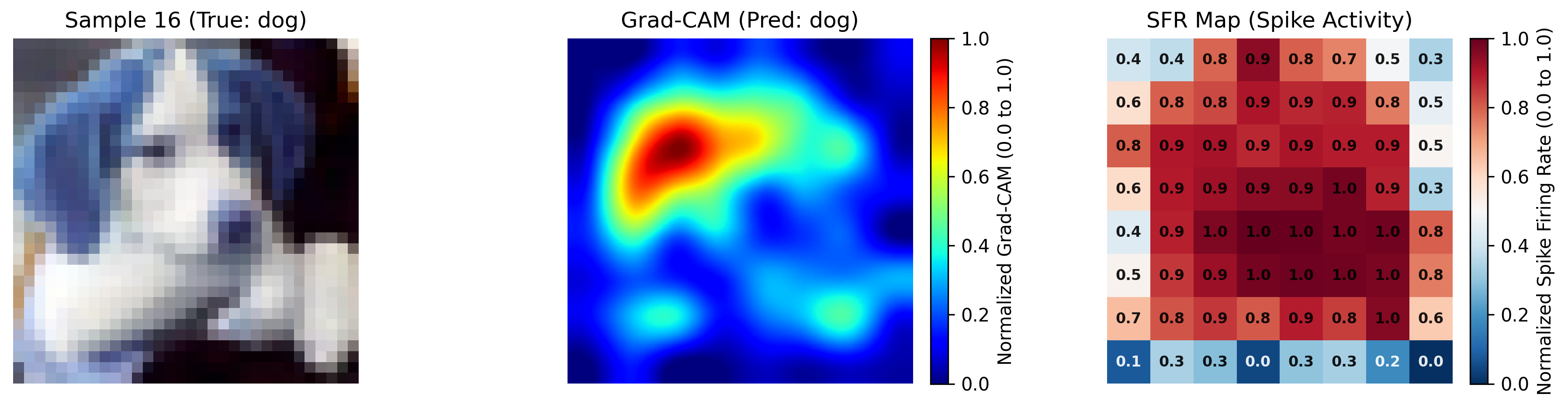}
        
        \vspace{1mm}

        % Row 5: sample 5
        \includegraphics[width=1.0\textwidth]{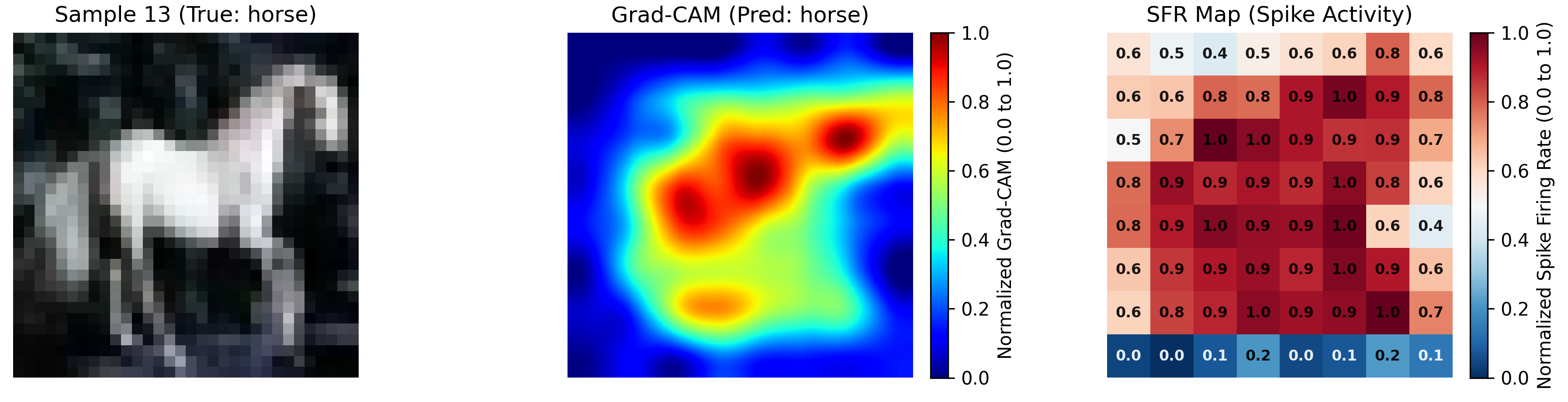}
      
    \end{minipage}
    \caption{The figure displays the visualization results for five different classes, each shown with three components from left to right: (1) the original input image from CIFAR10, (2) the Spiking Grad-CAM ($S^2TDPT$) visualization, and (3) the Spike Firing Rate (SFR) Map.
    The Grad-CAM heatmaps consistently exhibit sparse, localized activations concentrated around the target objects (e.g., the car body, dog's head, or horse's torso). This confirms that the model’s classification decision, derived from the features of the final Transformer block, is based on discriminative and task-relevant object features.}
    \label{F4}
\end{figure}

 \noindent Moreover, the proposed \textbf{$S^{2}TDPT$} model demonstrates notably superior energy efficiency compared to existing architectures. Based on the measurements reported in Table~\ref{T4}, $S^{2}TDPT$ reduces energy consumption by 15.5\% relative to SAFormer~\cite{SAFormer}, by 20.67\% compared to S-Transformer~\cite{SAFormer,spikedriventransformer}, and by 37.97\% with respect to Spikformer~\cite{Spikformer}. In comparison to the ANN Transformer baseline~\cite{r1}, the reduction reaches as high as 88.47\%. These results clearly indicate that $S^{2}TDPT$ achieves strong energy-aware performance while maintaining competitive classification accuracy.

\section{Discussion}

This study demonstrates that spike-timing–dependent plasticity (STDP) can provide a practical and biologically grounded alternative to dot-product operations in Transformer attention. By encoding token relevance through relative spike timing, the model eliminates multiplications and performs in-memory computation of attention weight matrices, while maintaining competitive performance at significantly lower energy cost. The qualitative alignment between firing-rate maps and gradient-based saliency indicates that the attention mechanism is inherently object-centered and internally interpretable. 

The proposed Spiking STDP Transformer achieves promising results on CIFAR-10 and CIFAR-100, and can be expanded on other datasets such as ImageNet \cite{imagenet}, with its 1,000 object categories, and specialized neuromorphic datasets like DVS-CIFAR10 \cite{cifar10dvs}. Currently, the architecture uses multi-step Leaky-Integrate-and-Fire (LIF) neurons, though alternative spiking neuron models such as Conductance-based LIF (CLIF), Gating LIF(GLIF), Kernel-based LIF(KLIF), and Parametric LIF( PLIF) can be extended for increasing accuracy and energy efficiency \cite{STEP}. 
We anticipate that Further refinement, including careful hyperparameter tuning, could enhance performance to state-of-the-art results.

Currently, training uses backpropagation with quantization on GPUs, but neuromorphic-native learning via on-device STDP could enable energy-efficient, fully online adaptation \cite{d1}. A recent study on biologically inspired spiking language models shows that they achieve GPT-2-level performance while maintaining interpretability~\cite{g1}. Beyond algorithmic advances, practical implementation on neuromorphic hardware is an essential step for validation. Recent advances in neuromorphic nano-materials have demonstrated that various 2D materials and metal-oxide systems and CMOS circuits exhibit robust STDP behavior suitable for implementing this brain-inspired attention mechanisms. Molecular memristors \cite{r66} exhibit both Hebbian and anti-Hebbian spike-timing-dependent plasticity (STDP) characteristics, with conductance changes spanning four orders of magnitude (\(200~\text{nS}\) to \(5.9~\text{mS}\)) \cite{r66}. Conducting polymer dendritic interconnections grown via electropolymerization emulate biological dendrite morphogenesis, enables tunable Hebbian plasticity \cite{r62,r64}. ${WS}_{2}$ memristors successfully replicate both symmetric and anti-symmetric Hebbian and anti-Hebbian STDP rules~\cite{r63}. Percolating Ag-islands and nanowires also shows synaptic behaviour for neuromrophic computing \cite{y3,y4}. Even standard bulk-silicon CMOS transistors, can exhibit neuronal and synaptic behaviors \cite{r65}. Recent work shows in-memory computing, memristor crossbars, and phase-change memory can drastically cut transformer attention energy while preserving accuracy~\cite{g2,g3,g4}. Thus, several neuromorphic materials and devices can potentially emulate the characteristics of an attention layer proposed in this work.

\section{Conclusion}

This work presents the Spiking STDP Transformer (S²TDPT), a foundational contribution toward truly neuromorphic Transformer architectures that integrate attention computation with the physics of synaptic plasticity. By reimagining the self-attention mechanism through spike-timing-dependent plasticity (STDP) and temporal coding, we address three critical limitations of existing spiking Transformers: the incompatibility of dot-product similarity with binary spike representations, the explicit materialization of attention score matrices that violates in-memory computing principles, and the lack of biological plausibility in attention computation.

The proposed approach replaces conventional query-key similarity calculations with an STDP-driven mechanism that encodes relevance through precise spike timing. This design naturally produces addition-only operations throughout the attention module, eliminating the need for softmax normalization operations. Critically, similarity scores are embedded directly within synaptic weights through local plasticity dynamics rather than stored as separate external matrices, achieving true co-localized memory-computation-a hallmark of efficient neuromorphic systems.

We have demonstrated the performance of our proposed architecture on CIFAR-10 and CIFAR-100 dataset. With 94.35\% and 78.08\% top-1 accuracies respectively and energy consumption of 0.49 mJ, S²TDPT surpasses multiple state-of-the-art spiking and traditional architectures while achieving exceptional energy efficiency gains (37.97\% reduction over Spikformer, 88.47\% over standard Transformers). Spiking Grad-CAM and spike-firing-rate maps show that the model learns object-centric, discriminative features, thereby validating the effectiveness of the biologically inspired attention mechanism, demonstrating its interpretability and explainable AI

Beyond empirical performance, this work represents a philosophical return to the foundational principles of neuromorphic engineering-moving beyond benchmark-driven optimization toward architectures grounded in neuroscience. The successful implementation of plasticity-driven attention within standard Leaky-Integrate-and-Fire neurons, combined with the emerging availability of STDP-capable device technologies demonstrates the practical feasibility of realizing Spiking STDP Transformer (S²TDPT) on next-generation neuromorphic hardware. This work may help us better understand and implement attention, while opening new research directions in neuroscience, neuromorphic devices, and artificial intelligence.

\section{Acknowledgment}

The authors acknowledge IIT Roorkee for the institute-funded SPARK Fellowship. Ankush Kumar acknowledges the Department of Science and Technology, India, for funding and the Faculty Initiation Grant at IIT Roorkee.

\appendix
\section{Appendix}
\subsection{Post-Encoder Feature Aggregation}
\label{A1}

The final layer of the Spiking Transformer encoder, $U_L$, produces a membrane potential tensor $U_L \in \mathbb{R}^{T \times N \times D}$, where $T$ is the number of timesteps, $N$ is the number of spatial patch tokens, and $D$ is the feature dimensionality. To transform this dynamic spatio-temporal representation into a fixed-length vector suitable for the classification head, a two-stage global pooling mechanism is applied: Global Temporal Mean Pooling (GTMP) followed by Global Average Pooling (GAP).

\textbf{Global Temporal Mean Pooling (GTMP):}
GTMP performs temporal integration by computing the mean activity of the membrane potential across all $T$ timesteps.
\begin{equation}
x_{\text{temp}} = \frac{1}{T} \sum_{t=1}^{T} U_L(t, \cdot, \cdot), \quad x_{\text{temp}} \in \mathbb{R}^{N \times D}
\end{equation}
where $U_L(t, \cdot, \cdot)$ denotes the entire $N \times D$ feature matrix at timestep $t$. The two dots ($\cdot$) indicate that the $N$ patch tokens and $D$ feature channels are preserved. The summation $\sum_{t=1}^{T}$ performs an element-wise average across the temporal dimension. GTMP functions as a temporal rate decoder. By averaging the accumulated membrane potential over all timesteps, it provides a stable measure of sustained spike rate and integrated activation for each feature and patch, effectively collapsing the $T$ dimension.

\textbf{Global Average Pooling (GAP):}
\textbf{Function:} GAP performs spatial aggregation by computing the mean of the time-integrated features across all $N$ spatial patch tokens.
\begin{equation}
x_{\text{final}} = \text{GAP}(x_{\text{temp}}) = \frac{1}{N} \sum_{i=1}^{N} x_{\text{temp}}(i, \cdot), \quad x_{\text{final}} \in \mathbb{R}^{D}
\end{equation}
The index $i$ ranges from $1$ to $N$, representing each spatial patch token. The single dot ($\cdot$) indicates that the $D$ feature dimensions are preserved. This operation averages the features across all $N$ tokens. GAP collapses the spatial dimension to yield a final $D$-dimensional global feature descriptor $x_{\text{final}}$. This operation acts as a structural regularizer, promoting robustness and translational invariance by ensuring that the classification decision relies on the average content across the entire input, rather than focusing on any localized feature.

\subsection{Memory Bottleneck in Attention Mechanisms}
\label{A2}
Traditional transformer architectures suffer from a fundamental memory bottleneck during attention computation, which arises from the explicit materialization of the intermediate attention score matrix $QK^{T}$. This matrix, with dimensions $N \times N$ where $N$ denotes the sequence length, scales quadratically and must be stored in memory before subsequent operations. Each element of this matrix is typically represented as a $32\text{-bit}~(4\text{-byte})$ floating-point number. As a result, the memory footprint increases rapidly with longer sequences. For instance, when $N = 512$, the attention score matrix occupies approximately 1~MB of memory; for $N = 4096$, it expands to 64~MB-already exceeding the capacity of many on-chip caches; and for $N = 16384$, it reaches roughly 1~GB, forcing repeated and energy-expensive main memory access. This intermediate matrix must be explicitly computed, stored, and accessed multiple times-first for normalization (softmax) and subsequently for multiplication with the value matrix $V$-leading to substantial memory bandwidth consumption and limiting scalability.

The other spiking mechanism employed in spiking transformer architectures alleviates this issue by leveraging sparse and low-precision spike-based representations. Specifically, the query ($Q$), key ($K$), Value ($V$) is represented as a binary spike signal (1~bit). Furthermore, attention scores are computed using addition-only operations, eliminating the need for multiply-accumulate (MAC) operations, and the softmax normalization step is removed, thereby reducing intermediate storage requirements. For example, with $N=512$ and feature dimension $D=64$, a traditional transformer requires approximately 1~MB for attention scores, and an additional 1~MB for softmax normalization and overhead. In contrast, the spiking transformer requires no additional softmax overhead, representing an estimated 20--30\% reduction in memory bandwidth requirements. The quadratic intermediate matrix still requires explicit storage and repeated memory access. 

Beyond these representational optimizations, the proposed STDP-based in-memory computing approach fundamentally departs from conventional transformer paradigms. Instead of computing and storing an explicit $N \times N$ attention weight matrix, the similarity between query and key spike patterns is encoded implicitly through Spike-Timing-Dependent Plasticity (STDP) dynamics. These local synaptic updates adjust connection weights based on relative spike timing, thereby embedding the $Q$–$K$ similarity directly within the synaptic state. In this framework, the attention score is not maintained as an external computational variable but is inherently stored and computed within the synapse itself, achieving true co-location of computation and memory(in-memory computing)-an essential characteristic of neuromorphic computing.

\end{document}